\documentclass[final]{cvpr}

\usepackage{times}
\usepackage{graphicx}
\usepackage{amsmath}
\usepackage{amssymb}
\usepackage{multirow}
\usepackage{array}  %
\usepackage{color}
\usepackage{animate}
\usepackage{tabularx}
\usepackage{placeins}
\usepackage[tight,scriptsize,sf,SF]{subfigure}

\def\R{\mathbb{R}}

\usepackage[pagebackref=true,breaklinks=true,colorlinks,bookmarks=false]{hyperref}

\begin{document}
\newcommand{\dbname}{MUSES}
% \newcommand
%%%%%%%%% TITLE
\title{
Multi-shot Temporal Event Localization: a Benchmark
}

\author{
Xiaolong Liu$^{1}$\thanks{Work done during an internship at Alibaba Group.}~~~~
Yao Hu$^2$~~~~
Song Bai$^{2,3}$\thanks{Corresponding author}~~~~
Fei Ding$^2$~~~~
Xiang Bai$^1$~~~~
Philip H.S. Torr$^3$\\
$^1$Huazhong University of Science and Technology\\
$^2$Alibaba Group~~~~
$^3$ University of Oxford\\
{\tt\small \{liuxl, xbai\}@hust.edu.cn, songbai.site@gmail.com} \\
{\tt\small \{feifei.df, yaoohu\}@alibaba-inc.com, philip.torr@eng.ox.ac.uk}
}

\maketitle
\thispagestyle{empty}
\pagestyle{empty}
%%%%%%%%% ABSTRACT
\begin{abstract}
Current developments in temporal event or action localization usually target actions captured by a single camera. However, extensive events or actions in the wild may be captured as a sequence of shots by multiple cameras at different positions.
In this paper, we propose a new and challenging task called multi-shot temporal event localization, and accordingly, collect a large-scale dataset called MUlti-Shot EventS (\dbname).~\dbname~has 31,477 event instances for a total of 716 video hours. The core nature of MUSES is the frequent shot cuts, for an average of 19 shots per instance and 176 shots per video, which induces large intra-instance variations. Our comprehensive evaluations show that the state-of-the-art method in temporal action localization only achieves an mAP of 13.1\% at IoU=0.5. As a minor contribution, we present a simple baseline approach for handling the intra-instance variations, which reports an mAP of 18.9\% on MUSES and 56.9\% 
on THUMOS14 at IoU=0.5. To facilitate research in this direction, we release the dataset and the project code at \url{https://songbai.site/muses/}.

% Current developments in temporal event or action localization usually target actions captured by a single camera. However, extensive events or actions in the wild may be captured as a sequence of shots by multiple cameras at different positions. In this paper, we propose a new and challenging task called multi-shot temporal event localization, and accordingly, collect a large-scale dataset called MUlti-Shot EventS (MUSES). MUSES has 31,477 event instances for a total of 716 video hours. The core nature of MUSES is the frequent shot cuts, for an average of 19 shots per instance and 176 shots per video, which induces large intra-instance variations. Our comprehensive evaluations show that the state-of-the-art method in temporal action localization only achieves an mAP of 13.1% at IoU=0.5. As a minor contribution, we present a simple baseline approach for handling the intra-instance variations, which reports an mAP of 18.9% on MUSES and 56.9% on THUMOS14 at IoU=0.5. To facilitate research in this direction, we release the dataset and the project code at \url{https://songbai.site/muses/}.
   
\end{abstract}

%%%%%%%%% BODY TEXT
\section{Introduction}
Driven by the increasing number of videos generated, shared and consumed every day, video understanding has attracted greater attention in computer vision especially in recent years. As one of the pillars in video understanding, temporal event (or action) localization~\cite{chao2018rethinking,long2019gaussian, shou2016temporal,xu2020g,yuan2017temporal,zeng2019graph,zhao2020bottom,zhao2017temporal} is a challenging task that aims to predict the semantic label of an action, and in the meantime, locate its start time and end time in a long video. Automating this process is of great importance for many applications,~\eg,~security surveillance, home care, human-computer interaction, and sports analysis.

\begin{figure}[tb]
\centering
\subfigure[]{
\animategraphics[width=0.22\textwidth,height=0.8in]{3}{imgs/animation/direct/img_000}{2899}{2906}
\label{fig:cut:direct}
}
\subfigure[]{
\animategraphics[width=0.22\textwidth,height=0.8in]{3}{imgs/animation/dissolve/img_00}{10209}{10219}
\label{fig:cut:dissolve}
}
\subfigure[]
{
\animategraphics[width=0.22\textwidth,height=0.8in]{3}{imgs/animation/cutin/img_0000}{114}{121}
\label{fig:cut:cutin}
}
% \subfigure[]
% {
% \animategraphics[width=0.22\textwidth,height=0.8in]{3}{imgs/animation/cutout/img_00}{21180}{21187}
% \label{fig:cut:cutout}
% }
\subfigure[]
{
\animategraphics[width=0.22\textwidth,height=0.8in]{3}{imgs/animation/crosscut/img_000}{8910}{8922}
\label{fig:cut:crosscut}
}
\caption{Different types of shot cuts, such as cutting on action (a), dissolve (b), cut-in (c), and cross-cut (d). \textbf{\textcolor{red}{Clickable}}: best viewed with Adobe Acrobat Reader; click to watch the animation.
}
\label{fig:example}
\end{figure}

\begin{figure*}[tb]
\centering
\includegraphics[width=0.95\linewidth]{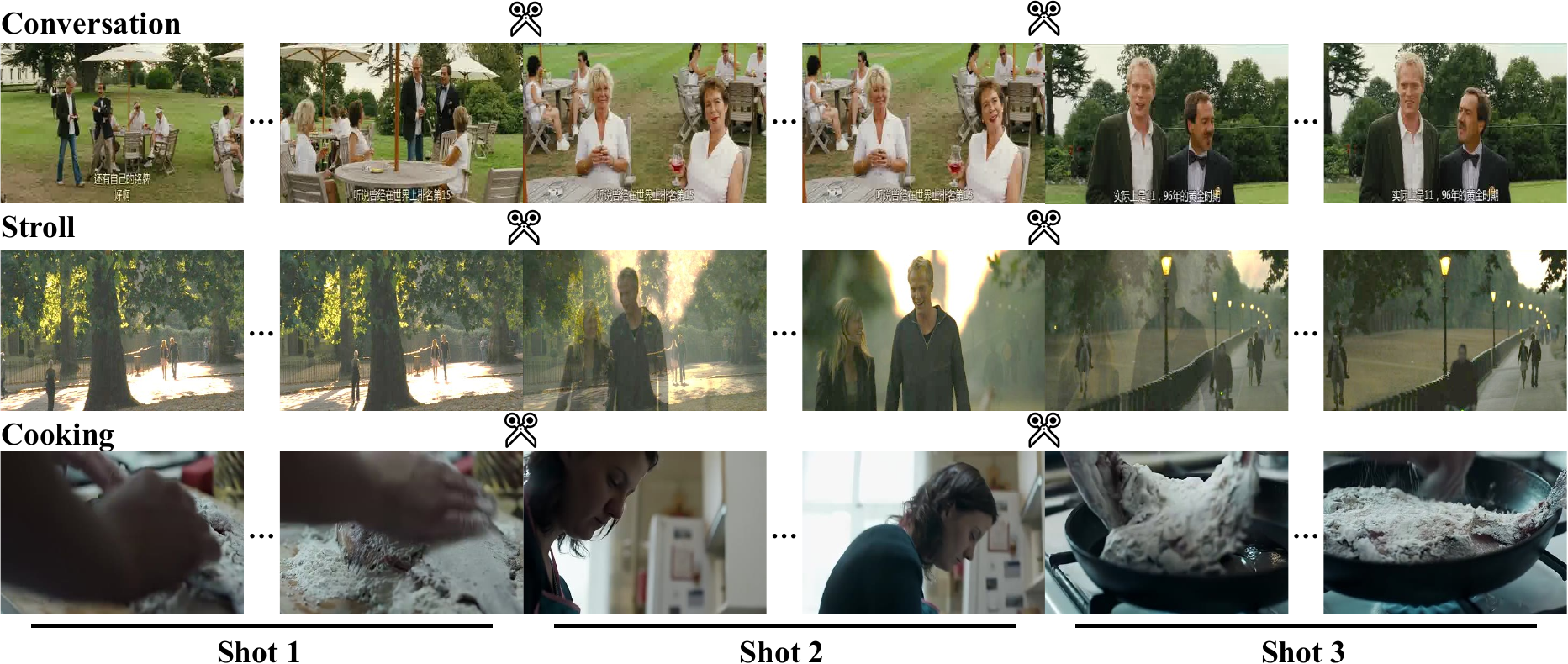}
\caption{
Examples of multi-shot events. In each row, we show three consecutive shots in an instance and select two frames per shot for illustration. The scissor icons indicate the shot boundaries.}
\label{fig:intro}
\end{figure*}

While fruitful progress has been made in this field, our work focuses on a real-world scenario that has not been seriously treated so far, that is, localizing events in TV shows and movies. We cast it as a new research topic named multi-shot temporal event localization, which is detailed below:

\textbf{1)~Motivation}:~Our work is motivated by some commercial applications of multi-shot temporal event localization, where truncated videos are automatically generated from TV shows and movies. For example, a trailer is generated for presenting the highlight to tease the audience, a video summary is generated to help the viewer grasp the important plot points and characters, or a mashup is generated for blending multiple clips of the same theme. To achieve this, a basic but time-consuming step is to find the materials,~\ie,~the video segments of interest, while the development of multi-shot temporal event localization will enable efficient material extraction and greatly improve the productivity of video content generation.

\textbf{2)~Characteristic}:~Compared with user-generated videos or surveillance videos, a unique characteristic of TV shows and movies is the highly frequent shot cuts. Herein, a shot means a single sequence of video frames taken by one camera without interruption\footnote{Note that the meaning of ``shot" in our work is different from those in one-shot/zero-shot/few-shot learning~\cite{fei2006one}.}. Because of the use of multi-camera shooting and professional editing techniques, a complete action or event in such videos is usually expressed as a sequence of meaningful short shots connected by cuts of various types, such as cutting on action,  dissolve, cut-in, and cross-cut (see Fig.~\ref{fig:example} for examples). In other words, the termination of a shot does not mean the end of the corresponding event or action.

\textbf{3)~Challenge}:~The key challenge of localizing events in TV shows and movies is the large intra-instance variation, induced by the nature of shot cuts. As can be observed in Fig.~\ref{fig:intro}, the view angles and the depth of fields across shots vary dramatically. Meanwhile, due to the existence of shot cuts, some side effects occur, such as scene change, actor change, and heavy occlusions. With such large variations within a single instance, the difficulty of localizing a complete event across shots is significantly increased.

\textbf{4)~Our Contribution}:~To facilitate the study of multi-shot temporal event localization, the main contribution of this work is a collection of a large-scale dataset named \textbf{MU}lti-\textbf{S}hot \textbf{E}vent\textbf{S} (MUSES). Unlike existing datasets (\eg,~THUMOS14~\cite{jiang2014thumos}, ActivityNet-1.3~\cite{caba2015activitynet}, HACS Segments~\cite{zhao2019hacs}) that are mainly built upon user-generated videos, the data source of MUSES is drama videos processed by professional editing techniques of the entertainment industry. Each instance is defined and annotated as an individual event that may take place across multiple shots. As a consequence, the number of shots in MUSES is as great as 19 per instance and 176 per video. MUSES is also a large-scale dataset that is suitable for training deep learning models, which contains 31,477 instances for a total of 716 video hours. 

We conduct a comprehensive evaluation of state-of-the-art methods on MUSES, including P-GCN~\cite{zeng2019graph}, G-TAD~\cite{xu2017r} and MR~\cite{zhao2020bottom}.
The results show that current methods cannot well handle the large intra-instance variations. The best performance is achieved by P-GCN~\cite{zeng2019graph}, which is 13.1\% mAP at IoU=0.5. It reveals the difficulty of our dataset in temporal event localization, and in the meantime, dramatically exposes the necessity of exploring methods specifically focusing on capturing intra-instance variations. As a minor contribution of this work, we present a simple baseline approach for multi-shot temporal event localization. The proposed baseline achieves an mAP of 18.9\% on MUSES and 56.9\% on THUMOS14~\cite{jiang2014thumos} at IoU=0.5. The dataset and the project code are publicly available for fellow researchers.

%---------------------------------------------------------- --------------
\section{Related Work}
Our work targets temporal event or action localization by contributing a new benchmark dataset. Existing datasets are mainly built upon user-generated videos, where less professional editing is involved. For example, THUMOS14~\cite{jiang2014thumos} focuses on sports events. ActivityNet-1.3~\cite{caba2015activitynet} extends the taxonomy from sports to human daily activities and significantly increases the number of categories and samples. HACS Segments~\cite{zhao2019hacs} shares the same lexicon as ActivityNet and further increases the size. In comparison, our dataset is based on drama videos processed by professional editing with frequent shot cuts, so the intra-instance variances are much greater.

\begin{table*}[tb]
\small
\centering
\begin{tabular}{|p{3.1cm}|*{4}{p{1.7cm}<{\centering}}cp{1.6cm}<{\centering}|}
\hline
Datasets & \#Videos & \#Categories & \#Instances & \#Total Hours & \#Categories per Video & Multi-shot  \\
\hline
\hline
THUMOS14~\cite{jiang2014thumos} & 413 &20 &6,365 &30 & 1.2 & -  \\
ActivityNet-1.3~\cite{caba2015activitynet}& 19,994&200&30,971&648 & 1 & - \\
HACS Segments~\cite{zhao2019hacs}& 50k&200 &139k& 2,048 & 1 & - \\
\dbname{} (ours)& 3,697 &25&31,477&716 & 3.3 & $\surd$ \\
\hline
\end{tabular}
\caption{Comparing MUSES with existing datasets for temporal event localization.}
\label{tab:stats}
\end{table*}

Previous methods on temporal event localization can be roughly categorized into two groups,~\ie,~\textbf{two-stage methods}~\cite{caba2016fast, heilbron2017scc,ni2016progressively,liu2020self,richard2016temporal,yuan2016temporal, zhao2017temporal} and \textbf{one-stage methods}~\cite{alwassel2018action,lea2017temporal, lin2017single, long2019gaussian,Ma_2016_CVPR, shou2017cdc,yuan2017temporal}, according to whether a standalone proposal generation stage is used.
For \textbf{two-stage methods}, we first generate a set of proposals,~\ie,~temporal segments that may contain an event, then leverage a classifier for semantic prediction. Proposal generations can be fulfilled by scoring handcrafted anchors~\cite{buch2017sst, chao2018rethinking, escorcia2016daps, gao2017turn, shou2016temporal,xu2017r}, grouping potential event boundaries~\cite{lin2018bsn, shou2017cdc,zhao2020bottom, zhao2017temporal}, or a combination of both~\cite{gao2018ctap,liu2019multi}. For example, Shou~\etal~\cite{shou2016temporal} sample regularly distributed segments as proposals and train a binary classifier to remove background segments.
% Zhao~\etal~\cite{zhao2017temporal} predict the probability of containing an action for each frame and generate proposals by grouping frames with high probabilities.
Lin~\etal~\cite{lin2018bsn} locate action boundaries by per-frame classification and group pairs of potential boundaries as proposals. To classify the proposals, Convolutional Neural Networks (CNNs)~\cite{gao2017turn, shou2016temporal, zhao2017temporal}, Recurrent Neural Networks (RNNs)~\cite{yuan2016temporal} or Graph Convolutional Networks (GCNs)~\cite{zeng2019graph} are usually used. For instance, Shou~\etal~\cite{shou2016temporal} employ the C3D~\cite{tran2015learning} network for proposal classification. Zeng~\etal~\cite{zeng2019graph} model proposal-proposal relations with GCNs to extract contextual information.
In \textbf{one-stage methods}, action localization can be implemented by classifying each pre-defined anchor~\cite{buch2017end,lea2017temporal, lin2017single,long2019gaussian} or each frame~\cite{Ma_2016_CVPR, shou2017cdc, yuan2017temporal}. For example, Lin~\etal~\cite{lin2017single} predict the classes and boundaries of all anchors simultaneously. Meanwhile, reinforcement learning is also exploited~\cite{yeung2016end}. Some methods~\cite{liu2019completeness,nguyen2018weakly, paul2018w,shou2018autoloc, wang2017untrimmednets,yu2019temporal} also try to learn action localization models in a weakly supervised manner, where only video-level labels are accessible.

This task is also related to action recognition~\cite{feichtenhofer2019slowfast,feichtenhofer2016spatiotemporal,lin2019tsm,qiu2017learning,simonyan2014two,tran2018closer,wang2013action, wang2016temporal, xie2018rethinking,yang2020temporal}, spatio-temporal action localization~\cite{ hou2017tube,kalogeiton2017action,li2020actions,singh2017online,wu2020context,yang2019step} and temporal action segmentation~\cite{ding2018weakly,lea2017temporal, lea2016segmental,lei2018temporal,li2020ms,richard2017weakly, wang2020boundary}. Action recognition aims to classify action in a given video, which is usually trimmed to include  one action only without background. Representative datasets include KTH~\cite{kth/schuldt2004recognizing}, Weizemann~\cite{weizmann/gorelick2007actions}, HMDB51~\cite{DBLP:conf/iccv/KuehneJGPS11}, UCF101~\cite{khurram2012ucf}, Sports-1M~\cite{karpathy2014large}, Charades~\cite{sigurdsson2016hollywood}, Kinetics~\cite{carreira2017quo}, Moments in Time~\cite{monfort2019moments}, Something-Something~\cite{goyal2017something}, EventNet~\cite{ye2015eventnet},
Youtube-8M~\cite{abu2016youtube} and FineGym~\cite{shao2020finegym}. Spatial-temporal action localization, which requires to locate both the temporal segment and spatial locations of actions. UCF101-24 \cite{khurram2012ucf}, JHMDB~\cite{jhuang2013towards} and AVA~\cite{gu2018ava} are often used for evaluation. 
Temporal action segmentation requires to label each frame in a long video by an action
class. Some popular datasets include 50Salads~\cite{stein2013combining}, GTEA~\cite{fathi2011learning} and Breakfast~\cite{kuehne2014language}, Hollywood Extended~\cite{bojanowski2014weakly}, and MPII Cooking 2~\cite{rohrbach2016recognizing}.

MUSES is also related to several movie datasets with action annotations, such as ThreadSafe~\cite{hoai2014thread}, AVA~\cite{gu2018ava} and MovieNet~\cite{huang2020movienet}, However, ThreadSafe and MovieNet focus on action classification and AVA is designed for spatio-temporal action localization respectively. Therefore, none of them is tailored for multi-shot temporal event localization.

\section{MUSES Dataset}
\label{sec:dataset}
The goal of this work is to build a large-scale dataset for temporal event localization, especially in the multi-shot scenario. In this section, we present the collection process, the statistic, the characteristic, the evaluation metric, and the baseline approach of the \dbname{} dataset.

\subsection{Data Collection}
\noindent \textbf{Category Selection.}~25 categories are chosen according to the occurrence frequency in drama, the difficulty of recognizing, and the audience appeal, including \textit{conversation, quarrel, crying, fight, drinking, eating, telephone conversation, horse riding, hugging, stroll, driving, chasing, gunfight, modern meeting, speech, ancient meeting, kissing, war, playing an instrument, dance, (human) flying, cooking, singing, bike riding}, and \textit{desk work}.
Some categories also appear in existing datasets (\eg,~ActivityNet-1.3~\cite{caba2015activitynet}), most of which are  daily-life actions such as \textit{drinking}. In addition, we include some events that are rare in daily life but common in dramas, such as \textit{gunfight}, and \textit{war}.

\vspace{1ex} \noindent \textbf{Data Collection.}~We begin with the drama database hosted by an online video sharing platform, where each drama is tagged with the genre, such as ``romance", ``action", ``crime", ``comedy", ``fantasy", ``war", ``life", ``sports", and ``science fiction". For each genre, we first exclude those dramas with a small number of views or of a low resolution, then randomly select dramas to ensure diversity. 
The 500 selected dramas contain 1,003 episodes that cover various years of production, genres, directors and actors, and also convey stories of different periodic eras. 

\vspace{1ex} \noindent \textbf{Annotation.}~The annotation of an instance is given as a set of the start time, the end time, and the category. Our annotation team is composed of 6 experienced annotators. The annotators are trained for a week before the formal annotation. To guarantee the quality, each video is labeled by at least two annotators. Two experts review the annotations provided by the annotators every week and resolve possible inconsistencies between different annotations. The whole annotation process lasts for 3 months.

Two noteworthy comments should be made here. First, as the motivation of this work reveals, an event may go through a series of shots, and sometimes the event is invisible or occluded during the shot cuts. For example, a shot of the audience may be inserted between two shots of a \textit{dance} event. We ask the annotation team to preserve the completeness of the \textit{dance} instance in this case.

Second, different instances may overlap temporally with each other. For example, \textit{telephone conversation} and \textit{driving} can happen simultaneously in a car. In this case, we ask the annotation team to extract two instances belonging to \textit{telephone conversation} and \textit{driving} separately.

\subsection{Statistics}
The dataset is split into two subsets, with 702 episodes for training and 301 for testing. The average length per episode is 42.8 minutes. Each episode is segmented into videos of around 10 minutes. After removing those that do not contain any events or actions, we obtain 3,697 videos, with 2,587 for training and 1,110 for testing. As a result,~\dbname~consists of 31,477 instances for a total of 716 video hours. 

\begin{figure}[tb]
\centering
\includegraphics[width=0.98\linewidth]{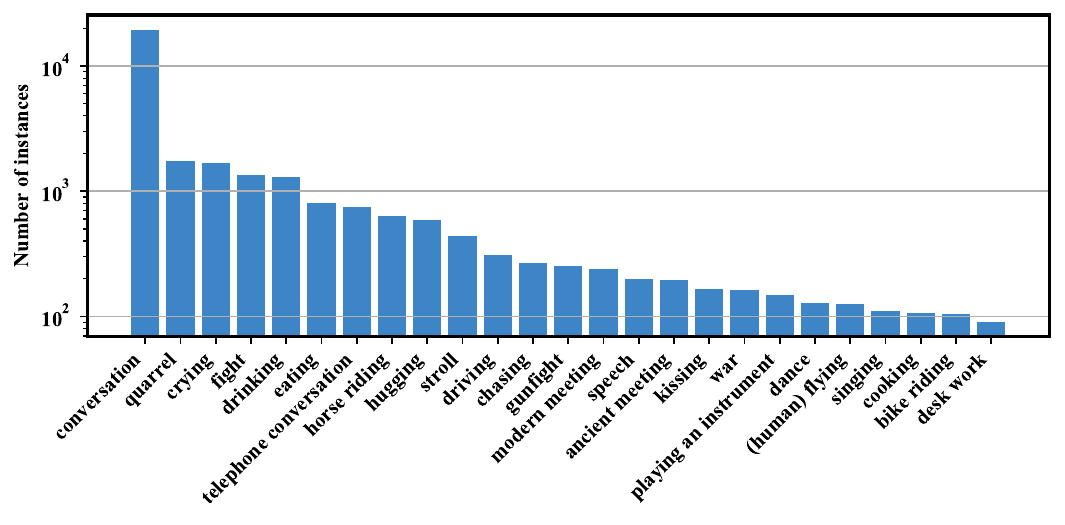}
\caption{Number of instances per category for \dbname{}. Note that the Y-axis is in logarithm scale.}
\label{fig:class_distribution}
\end{figure}

In Table~\ref{tab:stats}, we give a comparison of 
dataset statistics between MUSES and some representative datasets, including THUMOS14~\cite{jiang2014thumos}, ActivityNet-1.3~\cite{caba2015activitynet} and HACS Segments~\cite{zhao2019hacs}. In terms of scale, MUSES is slightly larger than ActivityNet-1.3 but smaller than HACS Segments. The length of videos varies from 300 to 1,151 seconds with an average of 698 seconds, which is about 3 times longer than THUMOS14. 

\begin{figure}[tb]
\centering
\subfigure[]
{
\includegraphics[width=0.22\textwidth]{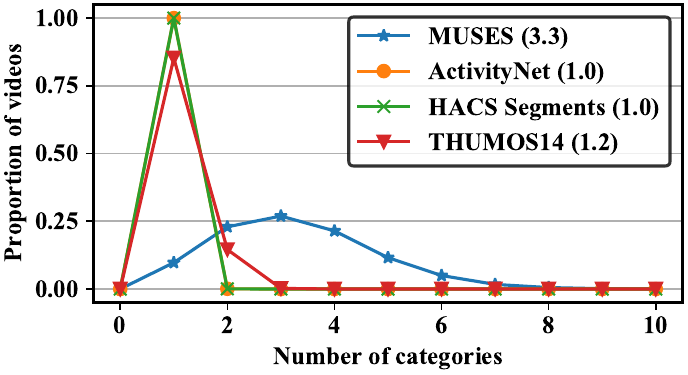}

\label{fig:classes_per_video}
}
\subfigure[]
{
\includegraphics[width=0.22\textwidth]{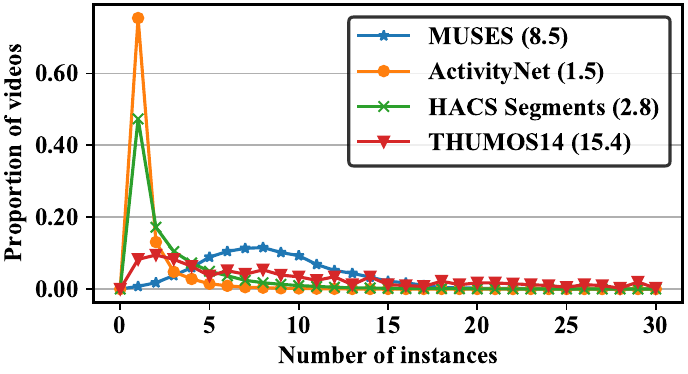}
\label{fig:instances_per_video}
}

\vspace{-1mm}
\subfigure[]
{
\includegraphics[width=0.22\textwidth]{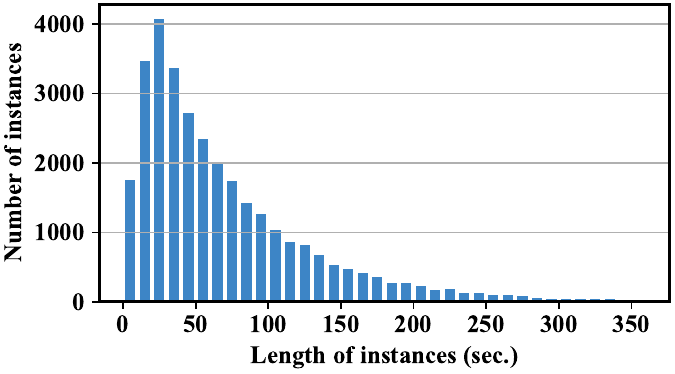}
\label{fig:inst_len}
}
\subfigure[]
{
\includegraphics[width=0.22\textwidth]{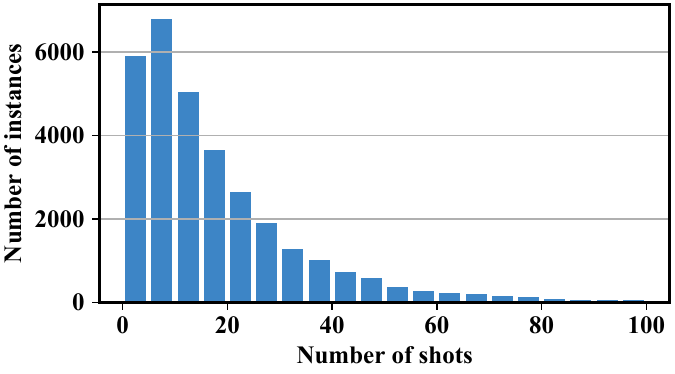}
\label{fig:shot_distribution}
}
\caption{Comparison of datasets: the number of categories (a) and instances per video (b). The numbers in the brackets are the average values of instances and categories per video. The distribution of instance length (c) and the number of shots (d) on MUSES.}
\label{fig:instance_density}
\end{figure}

Fig.~\ref{fig:class_distribution} depicts the category distribution over instances. As it shows, MUSES contains at least 90 instances per category and an average of 1,260 instances. We present the category distribution over videos in Fig.~\ref{fig:classes_per_video}, which suggests that most videos in MUSES contain instances from 3 categories while those in the other datasets are mainly from 1 semantic class. The instance distribution over videos is given in Fig.~\ref{fig:instances_per_video}. It can be observed that most videos in MUSES contain 5 to 10 instances with an average of 8.5 instances. Richer yet noisy contextual information is brought by the nature of multi-category and multi-instance in a single video, which encourages the development of advanced techniques for contextual reasoning. 
Fig.~\ref{fig:inst_len} presents the distribution of instance length, from which we find that most instances last for 20 to 40 seconds.

\begin{figure}[tb]
\centering
\includegraphics[width=\linewidth]{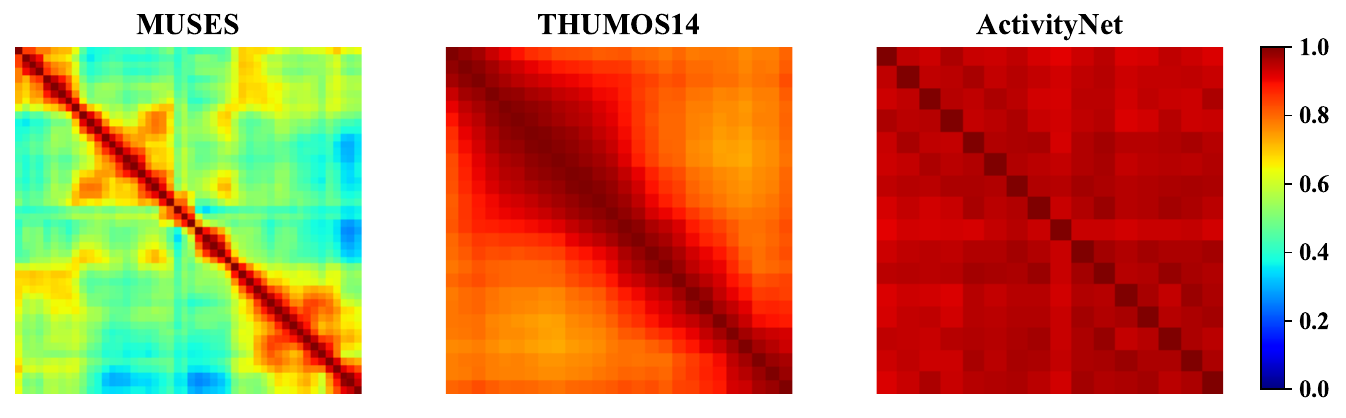}
% \fbox{\rule{0pt}{1in} \rule{0.9\linewidth}{0pt}}
\caption{Illustrations of self-similarity samples on MUSES, THUMOS14, and ActivityNet-1.3. The intra-instance variations on MUSES are much greater than that on the other datasets.}
\label{fig:self_similarity}
\end{figure}

Besides, it is found that 15.8\% of the instances overlap with another instance with an IoU$>$0.5. Some frequent pairs include \textit{playing an instrument} $\leftrightarrow$ \textit{singing}, \textit{dance} $\leftrightarrow$ \textit{singing}, \textit{kissing} $\leftrightarrow$ \textit{hugging}, \textit{crying} $\leftrightarrow$ \textit{hugging}, \textit{conversation} $\leftrightarrow$ \textit{eating}, \textit{war}  $\leftrightarrow$ \textit{horse riding}, \textit{fight} $\leftrightarrow$ \textit{(human) flying} $\leftrightarrow$ \textit{driving} and \textit{telephone conversation} $\leftrightarrow$ \textit{driving}. The reason behind is that those events or actions usually happen in the same scene.
%Some examples of instance overlapping are given in Fig.~\ref{fig:instance_overlapping}.

\subsection{\label{sec:characteristics} Characteristics}
The primary property of MUSES is frequent shot cuts inside a single instance.
Different from previous datasets~\cite{caba2015activitynet,jiang2014thumos,zhao2019hacs} for temporal event localization that mainly collect user-generated videos where less editing is involved, MUSES is built upon drama videos made by the industry. In these videos, professional editing techniques are widely used for different purposes, such as removing less informative segments or guiding the viewer's attention.
As a consequence, an event in such videos is usually expressed by a sequence of short shots connected by various types of shot cuts. Fig.~\ref{fig:shot_distribution} presents the distribution of the number of shots over instances. The average number of shots is 19 per instance and 176 per video. Frequent shot cuts induce large intra-instance variations, as we can see from Fig.~\ref{fig:intro} that different shots differ significantly in the view angles, depth of fields, actors, and backgrounds.
\begin{figure*}[tb]
\centering
\includegraphics[width=0.91\linewidth]{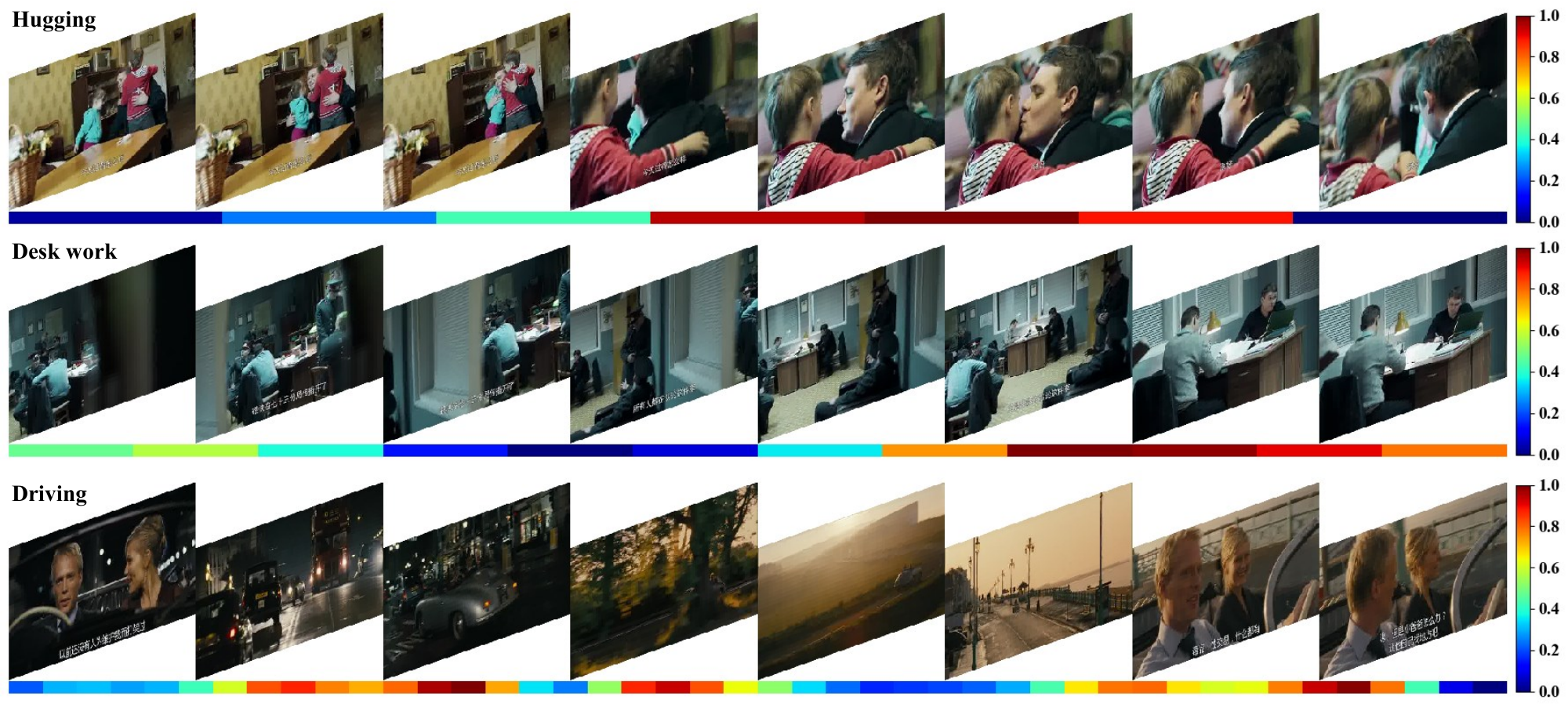}
\caption{Class activation mapping (CAM) of some ground truth instances in MUSES.}
\label{fig:cam}
\end{figure*}

The intra-instance variations can be further evidenced in the feature space. 1)~\textbf{Self-similarity}: we compute the cosine similarity of each pair of snippets of a single instance using the I3D~\cite{carreira2017quo} features, then report the average standard deviation of such self-similarity. The average standard deviation on MUSES is 0.16, while that on THUMOS14 and ActivityNet-1.3 is 0.01 and 0.09, respectively. It suggests that the self-similarity of instances on MUSES is generally smaller than that on the other datasets. 
Fig.~\ref{fig:self_similarity} further depicts some examples of self-similarities on different datasets. We find that in MUSES, the snippet pairs with high similarities usually occur in the same shot. 2) \textbf{Class activation mapping:} we leverage CAM~\cite{zhou2015cnnlocalization} to visualize the contribution of each temporal snippet to the final prediction. As shown in Fig.~\ref{fig:cam}, the individual contribution is significantly different, which again reveals the intra-instance variations. It also indicates that a considerable number of snippets are less discriminative, caused by the shot cuts, to a precise prediction of the semantic meaning and localization of the corresponding instance.

\subsection{Evaluation Metrics}
To evaluate the performance of multi-shot temporal event localization, we employ mean Average Precision (mAP). A prediction is considered as a true positive if its IoU with the ground truth is greater than a threshold $\alpha$ and the predicted category is correct, denoted as $\text{mAP}_{\alpha}$. $\alpha$ is set to 0.3 to 0.7 with a stride of 0.1 following the convention of THUMOS14. Meanwhile, we also report an average of mAPs, which is computed over the 5 IoU thresholds.

Moreover, to quantify the capability of algorithms in handling shot cuts, we further resort to mAPs with different numbers of shots. To be concrete, we divide all the instances into 3 groups according to the number of shots, and report $\text{mAP}_{\text{small}}$ (fewer than 10 shots), $\text{mAP}_{\text{medium}}$ (10 to 20 shots), and $\text{mAP}_{\text{large}}$ (no fewer than 20 shots). In MUSES, the 3 groups make up $39.8\%$, $27.5\%$, and $32.7\%$ of all instances, respectively.
\begin{figure*}[t]
    \centering
    \includegraphics[width=0.91\linewidth]{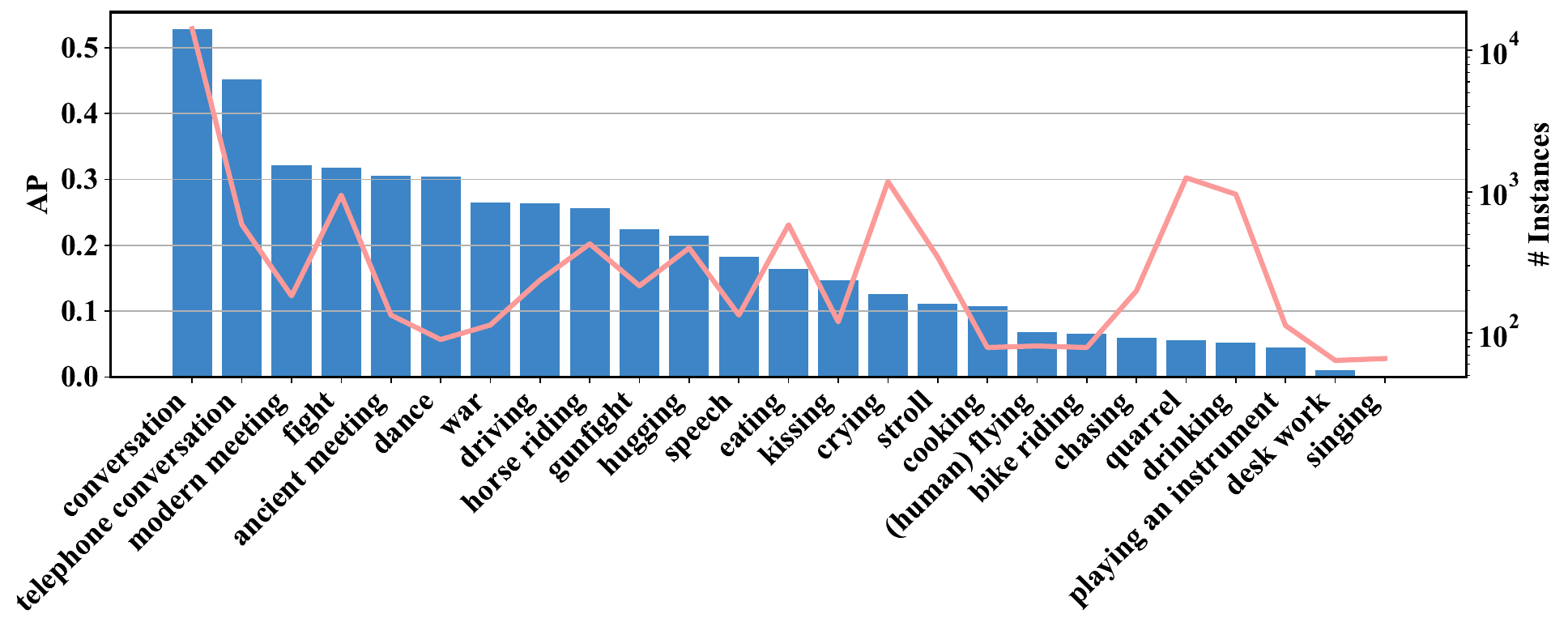}
    \caption{Event localization performance in terms of AP (bars) and the number of training instances per category (lines) on MUSES.}
    \label{fig:per_class_ap}
\end{figure*}

\subsection{Baseline Approach}

\noindent\textbf{Pipeline.}~We adopt the detection by classifying proposal paradigm following Fast R-CNN~\cite{girshick2015fast}. Taking a video and a set of temporal proposals (refer to the supplementary material for proposal generation) as input, our baseline approach is comprised of three steps: 1) feature extraction. As~\cite{zeng2019graph,chao2018rethinking, zhao2020bottom}, we use the I3D~\cite{carreira2017quo} network to extract a 1D feature for the input video; 2) temporal aggregation. The feature will be forwarded into a temporal aggregation module, which is detailed below, to mitigate the intra-instance variations caused by shot cuts; 3) proposal evaluation. For each proposal, a feature representation is extracted via RoI pooling \cite{girshick2015fast}. Two classifiers are utilized respectively to predict the category and the completeness of the proposal. A boundary regressor is also employed to adjust the boundaries.

\vspace{1ex}\noindent\textbf{Temporal Aggregation.}~As analyzed in Sec.~\ref{sec:characteristics}, shot cuts will result in large inter-snippet variations. One of the keys is to enhance the feature coherence within a single instance. To this end, we present a simple yet effective module called temporal aggregation to improve the feature discriminative power of each snippet.

Let $\mathbf{X}\in\R^{T\times C}$ be the video feature extracted from the I3D network, where $T$ and $C$ denote its length and the dimension respectively. We first evenly divide $\mathbf{X}$ into $H$ units of length $W$ as $\mathbf{X'}=\text{reshape}(\mathbf{X})\in\R^{H\times W\times C}$,
where $\mathbf{X'}_{ij}=\mathbf{X}_{iW+j}$ and $T=H\times W$. As a result, each row in $\mathbf{X'}$ correspond to consecutive snippets in the same unit, and each column in $\mathbf{X'}$ correspond to non-adjacent snippets with a stride of $W$ (a configurable parameter) in $\mathbf{X}$.

We then apply the standard 2D convolution to $\mathbf{X'}$ as $\mathbf{Y'}=\mathbf{W}*\mathbf{X'}$,
where $\mathbf{W}\in\R^{k_H \times k_W}$ is the convolution kernel.
In this way, we can obtain both short-term (within each row) and long-term (across multiple rows) information for each multi-shot instance, so that the feature variation is reduced significantly.
It is easy to find that the receptive field is equivalent to $(k_H-1)W+k_W$ and one can adjust the receptive field by tuning $W$ without changing the kernel size.

The output feature map $\mathbf{Y'}$ is re-casted into 1D view, as $\mathbf{Y}=\text{reshape}(\mathbf{Y'})\in\R^{T\times C}$, which is of the same size as the input feature $\mathbf{X}$. Therefore, the temporal aggregation module is  easy to implement using 2D convolutions and acts as a cheap plug-in. There are a few alternatives for modeling the long-term information, such as dilated 1D convolution, deformable 1D convolution~\cite{dai2017deformable},~\etc.~Our experiments suggest that they achieve inferior performance to our temporal aggregation module.

Moreover, as the duration of events and shots in MUSES vary significantly, we adopt a split-transform-merge strategy~\cite{xie2017aggregated} and build a multi-branch block to handle the scale changes. More precisely, we use $K$ temporal aggregation modules, each of which uses a convolution kernel of different sizes. The output of each branch is fused by summation. Thanks to the various kernel sizes, the learned feature integrates temporal information of different scales and is potentially capable of better dealing with events of different durations.

\section{Experiments}
In this section, we give a thorough study of the newly collected MUSES dataset. In addition, we report the results on THUMOS14~\cite{jiang2014thumos} and ActivityNet-1.3~\cite{caba2015activitynet} under the conventional experiment setup following \cite{shou2016temporal, zhao2017temporal, lin2018bsn, lin2019bmn}. Due to space limitation, we put the experimental details, such as the network architecture, the loss functions, the training strategy, some tricks like boundary extension~\cite{dai2017temporal} and the hyper-parameters, in the supplementary material. 
\begin{figure*}[tb]
\centering 
\subfigure[]
{
\includegraphics[width=0.55\textwidth]{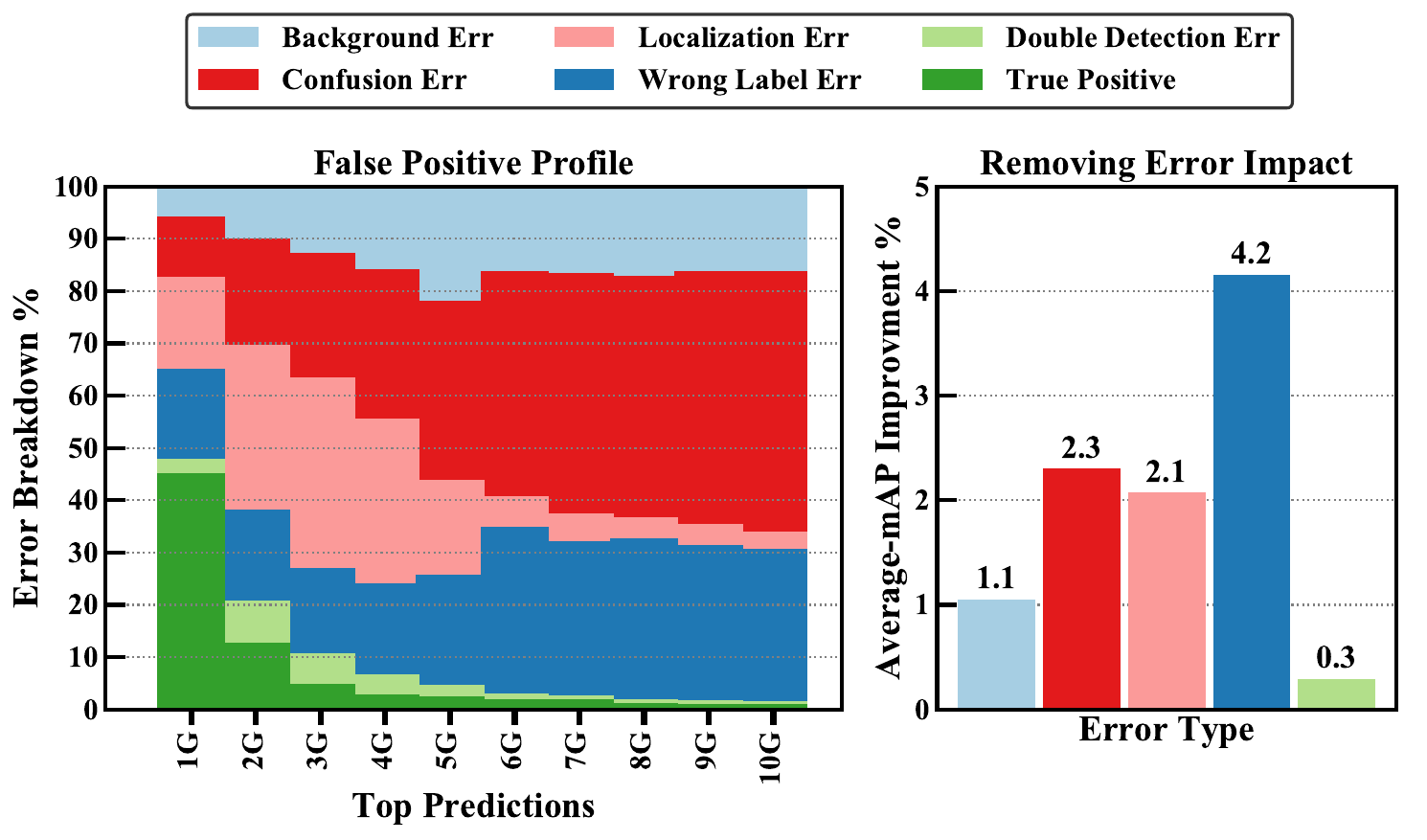}
\label{fig:error:fp}
}
\subfigure[]
{
\includegraphics[width=0.34\textwidth]{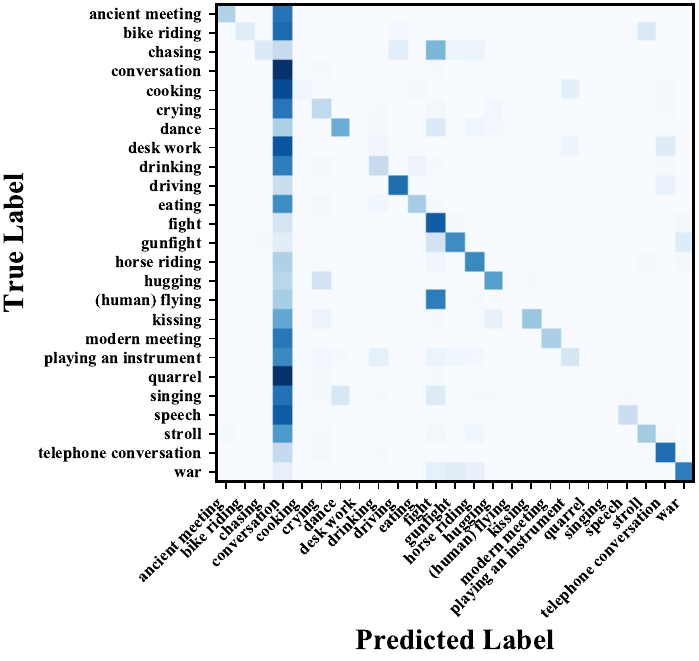}
\label{fig:error:cm}
}
\caption{Error analysis on MUSES. (a) \textbf{Left}: the error distribution over the number of predictions per video. G means the number of ground truth instances. (a) \textbf{Right}: the impact of error types, measured by the improvement gained from resolving a particular type of error. See Sec.~\ref{subsec:exp:discussions} for the definitions of the errors. (b) Confusion matrix.}
\label{fig:error} 
\end{figure*}

\subsection{Analysis of MUSES Dataset}
\label{subsec:exp:discussions}
\noindent \textbf{Which Categories Are More Challenging?}~We present the localization performance and the number of training instances per category in Fig.~\ref{fig:per_class_ap}. The best performance is witnessed in the \textit{conversation} category, probably owing to the largest number of training instances. Meanwhile, categories related to objects and scenes, such as \textit{telephone conversation} and \textit{modern meeting}, and categories that have salient motion features, such as \textit{fight} and \textit{dance}, are relatively easier to be coped with.
 The top-5 challenging categories are \textit{singing}, \textit{desk work}, \textit{playing an instrument}, \textit{drinking}, and \textit{quarrel}. The poor performance might be caused by less training data or larger diversity.

\vspace{1ex}\noindent\textbf{Error Analysis.}~Using the protocol devised by~\cite{alwassel2018diagnosing}, we present an error analysis on MUSES in Fig.~\ref{fig:error:fp}. According to its IoU with the matched ground truth (denoted as gIoU) and whether the predicted label is correct, a false positive can be divided into five types, including Double Detection Error (a duplicate prediction with $\text{gIoU}\geq\alpha$ and a correct label that is not the highest scoring one), Wrong Label Error ($\text{gIoU}\geq\alpha$, incorrect label), Localization Error ($0.1\leq\text{gIoU}<\alpha$, correct label), Confusion Error ($0.1\leq\text{gIoU}<\alpha$, incorrect label), and Background Error ($\text{gIoU}<0.1$).
 
As Fig.~\ref{fig:error:fp} left shows, Confusion Error, Localization Error, and Wrong Label Error are the three main sources of false positives. It indicates that incorrect boundaries and incorrect labels are responsible for false positive predictions, which are the side effect of frequent shot cuts. Different from the error distribution on ActivityNet-1.3 where Localization Error occurs most (quoted from ~\cite{alwassel2018diagnosing}), more Wrong Label Errors and Confusion Errors are observed on MUSES. This phenomenon suggests that the classification task is more difficult on MUSES compared with that on ActivityNet-1.3. Meanwhile, Fig.~\ref{fig:error:fp} right suggests that the performance boost is more remarkable if the classification task is well addressed.

To dissect Wrong Label Error, we present the confusion matrix in Fig.~\ref{fig:error:cm}. It is observed that instances in many categories are easy to be mis-classified into \textit{conversation}. The possible reasons are two folds: 1) 
% \textit{conversation} has the most training samples, so the learned model favors less the other minority classes; 
imbalanced number of training samples among different classes;
2) as one of the key visual cues in \textit{conversation}, the close-ups of characters also exist in the other categories, such as \textit{modern meeting}, \textit{quarrel}, \textit{speech}, \textit{desk work},~\etc.~Besides, event pairs with high concurrency, such as \textit{fight} $\leftrightarrow$ \textit{(human) flying} and \textit{singing} $\leftrightarrow$ \textit{dance}, are easy to confuse. 

\vspace{1ex}\noindent \textbf{Effect of Temporal Aggregation.}~In Table~\ref{tab:tvel_tam}, we compare the performance of different variants of our method. As can be observed, removing temporal aggregation (vanilla) leads to a decrease of $2.8\%$ in terms of mAP$_{0.5}$ on MUSES. Replacing temporal aggregation with dilated 1D convolution or deformable 1D convolution also leads to a performance drop. Meanwhile, an integration of multiple temporal aggregation modules at different scales outperforms its single-scale counterparts. The performance gain stems from the fact that temporal aggregation is capable of enhancing the feature coherence within each single instance to a certain extent. To verify this, we follow the gist described in Sec.~\ref{sec:characteristics} to compute the standard deviation of self-similarities, and find that the standard deviation is decreased from 0.16 to 0.09 after temporal aggregation is applied.

\begin{table}[tb]
\centering
\begin{tabular}{|l|ccc|}
\hline
Methods  & $k_H\times k_W$ & W & mAP$_{0.5}$ \\
\hline \hline
Vanilla & - & -  & 16.1 \\
Dilated Conv. & - & -  & 17.1 \\
Deformable Conv.~\cite{dai2017deformable} & - & -  & 17.0 \\
\hline
\hline
\multirow{4}{*}{Ours (Single-scale)} & 1$\times$ 3 & 3  & 16.0 \\
 & 3$\times$ 3 & 3  & 16.8 \\
 & 3$\times$ 3 & 6  & 17.5 \\
 & 3$\times$ 3 & 9  & 17.2 \\
\hline
\hline
Ours (Multi-scale) &- & -& \textbf{18.9} \\
\hline
\end{tabular}
\caption{Performance analysis of temporal aggregation in terms of mAP$_{0.5}$.}
\label{tab:tvel_tam}
\end{table}

\begin{table*}[tb]
\small
\centering
\begin{tabular}{|p{2.1cm}|*6{p{1.18cm}<{\centering}}|*3{p{1.2cm}<{\centering}}|}
\hline
Methods & mAP$_{0.3}$ & mAP$_{0.4}$ & mAP$_{0.5}$ & mAP$_{0.6}$ & mAP$_{0.7}$ & mAP & mAP$_{\text{small}}$    & mAP$_{\text{medium}}$    & mAP$_{\text{large}}$    \\ \hline
\hline
Random &1.20&0.64&0.29&0.10&0.03& 0.45        &0.02 & 0.06&0.41\\ 
MR~\cite{zhao2020bottom}&12.9&	11.3&	9.2&	7.6&	5.9&    9.4      &4.1&6.9&7.5\\
G-TAD~\cite{xu2020g}&19.1&	14.8&	11.1&	7.4&4.7&      11.4           &7.3&9.6&4.6\\
P-GCN~\cite{zeng2019graph} & 19.9& 17.1 &13.1& 9.7& 5.4&  13.0            &6.3 &8.7 &7.9\\
Ours & \textbf{25.9}&\textbf{22.6}&\textbf{18.9}&\textbf{15.0}&\textbf{10.6}&  \textbf{18.6}    & \textbf{7.4}&\textbf{11.9} &\textbf{17.3} \\
\hline
\end{tabular}
\caption{Performance evaluation of state-of-the-art methods on the newly collected \dbname{} dataset.}
\label{tab:tvel_sota}
\end{table*}

\vspace{1ex}\noindent \textbf{State-of-the-art Evaluations on MUSES.}~We present an evaluation of state-of-the-art methods whose code is publicly available, including P-GCN~\cite{zeng2019graph}, G-TAD~\cite{xu2020g}, and MR~\cite{zhao2020bottom}. Note that G-TAD and MR only focus on class-agnostic proposals, so we train an MLP on I3D features to generate category predictions. To ensure a fair comparison, the video features are kept the same for all the methods and the same proposals are used in P-GCN and our approach. 
As shown in Table~\ref{tab:tvel_sota}, our method achieves the highest mAPs among all the methods. However, the achievement is far below the state-of-the-art performance on THUMOS14 (an mAP of around 52\% at IoU=0.5, refer to~\cite{xu2020g, zeng2019graph}) and ActivityNet-1.3 (an mAP of around 50\% at IoU=0.5, refer to~\cite{lin2019bmn, xu2020g}), which reveals the difficulty of our dataset. 
% Note that events with more shots are often longer. Generally, longer events are easier to localize than shorter events. 
Note that mAP$_{\text{large}}$ can be higher than mAP$_{\text{medium}}$ and mAP$_{\text{small}}$, as events with more shots are often longer thus easier to localize. 
%This is similar to largesmall/ objects in object detection.

\subsection{Evaluations on THUMOS14 and ActivityNet}
Our baseline approach leverages temporal aggregation which is originally designed for handling intra-instance variations caused by frequent shot cuts. Here, we report the performance of temporal aggregation on THUMOS14 and ActivityNet-1.3.
Table~\ref{tab:thumos14_sota} presents the comparison with the state-of-the-art methods on THUMOS14. Our approach achieves the highest mAPs at different IoU thresholds. In particular, we outperform the second-best entry, that is G-TAD~\cite{xu2020g}, by an absolute improvement of 5.3\% in terms of mAP at IoU=0.5.
Table~\ref{tab:anet_sota} presents the performance comparison on ActivityNet-1.3. We build our model upon BMN~\cite{lin2019bmn}, one of the state-of-the-art algorithms with publicly available code. As it shows, integrating temporal aggregation with BMN is also useful and we report an 
%mAP of 50.02\% at IoU=0.5. 
average mAP of 33.99\%.

\begin{table}[tb]
\small
\centering
\begin{tabular}{|l|*5{p{0.78cm}<{\centering}}|}
\hline
% Methods& 0.3 & 0.4 & 0.5 & 0.6 & {0.7}\\ \hline \hline
Methods& mAP$_{0.3}$ & mAP$_{0.4}$ & mAP$_{0.5}$ & mAP$_{0.6}$ & mAP$_{0.7}$\\ \hline \hline
{SCNN~\cite{shou2016temporal}}         &  36.3 &   28.7   & 19.0  & 10.3  & 5.3  \\
{SMS~\cite{yuan2017temporal}}                              & 36.5 & 27.8 &   17.8   &   -   &    -  \\
{CDC~\cite{shou2017cdc}}                   & 41.3 & 30.7  & 24.7 &  14.3 &  8.8 \\
{Dai~\etal~\cite{dai2017temporal}}                & - & 33.3  & 25.6 &  15.9 &  9.0 \\
{R-C3D~\cite{xu2017r}}                      & 44.8 & 35.6  & 28.9 &  19.1 &  9.3 \\
{SS-TAD~\cite{buch2017end}}                           & 45.7 & -  & 29.2 &  - &  9.6 \\
{SSN~\cite{zhao2017temporal}}                          & 50.6 & 40.8  & 29.1 &  - &  - \\
{TAL-Net~\cite{chao2018rethinking}}         & 53.2 & 48.5 & 42.8 & 33.8 & 20.8 \\ 
{GTAN~\cite{long2019gaussian}} &57.8	&47.2	&38.8& -& -\\
{P-GCN~\cite{zeng2019graph}}    & 63.6 &57.8 &49.1 &  - &- \\
% {G-TAD~\cite{xu2020g}}  &54.5 & 47.6& 40.2& 30.8& 23.4 \\
{G-TAD~\cite{xu2020g}} & 66.4& 60.4& 51.6& 37.6& 22.9 \\
{MR~\cite{zhao2020bottom}}& 53.9 &50.7& 45.4& 38.0& 28.5 \\
{Ours } & \textbf{68.9} & \textbf{64.0} & \textbf{56.9} & \textbf{46.3} & \textbf{31.0} \\ \hline 
\end{tabular}
\caption{Performance comparison on THUMOS14 in terms of mAP at IoU thresholds from 0.3 to 0.7.}
\label{tab:thumos14_sota} 
\end{table}

\begin{table}[tb]
\small
\centering
\begin{tabular}{|p{2cm}|*4{p{1cm}<{\centering}}|}
\hline
Methods & mAP$_{0.5}$ &mAP$_{0.75}$ &mAP$_{0.95}$ &mAP\\ \hline \hline
{R-C3D~\cite{xu2017r}}&26.80 &- &- &12.70\\
{TAL-Net~\cite{chao2018rethinking}}&38.23&18.30&1.30&20.22\\
{GTAN~\cite{long2019gaussian}}& \textbf{52.61}&34.14&\textbf{8.91}&\textbf{34.31}\\
{BMN~\cite{lin2019bmn}}&50.07&34.78&8.29&33.85\\
{BMN*~\cite{lin2019bmn}}&49.66&33.85&7.86&33.45\\
{Ours}&50.02&\textbf{34.97}&6.57&33.99\\
\hline
\end{tabular}
\caption{Performance comparison on the validation set of ActivityNet-1.3 at different IoU thresholds. * indicates the result is our re-implementation with the publicly available code.}
\label{tab:anet_sota}
\end{table}
% }

\section{Conclusion}
Truncating TV shows and movies into concise and attractive short videos has been a popular way of increasing click-through rates in video sharing platforms, where localizing temporal segments of interest is the kick-off step. However, temporal event or action localization in such video sources is more or less ignored by our research community. And existing methods cannot well handle the intra-instance variations caused by the frequent shot cuts as TV shows and movies are usually post-processed by professional editing techniques. 

To enable automatic video content generation with efficient and scalable material extraction in TV shows and movies, we define a new task called multi-shot temporal event localization that aims to localize events or actions captured in multiple shots. A large-scale dataset, called MUSES, is collected for this study. Built upon drama videos, MUSES provide rich multi-shot instances with frequent shot cuts, which induces great intra-instance variations and brings new challenges to current approaches. A comprehensive evaluation shows
that the state-of-the-art methods in this field fail to cope with frequent shot cuts and reveal the difficulty of the MUSES dataset. MUSES could serve as a benchmark dataset and facilitate research in temporal event localization. 

MUSES will be updated in the future by adding multi-modal annotations, such as subtitles, audio, facial expression,~\etc~The setting of weakly-supervised multi-shot temporal event localization will be also defined and explored on top of MUSES. We hope that MUSES could trigger and advance the application of temporal event localization techniques to commercial products in video content generation.

\vspace{1ex}\noindent\textbf{Acknowledgments.}~This work was supported by National Key R\&D Program of China (No. 2018YFB1004600), to Dr. Xiang Bai by the National Program for Support of Top-notch Young Professionals, Alibaba Group through Alibaba Research Intern Program, Royal Academy of Engineering under the Research Chair and Senior Research Fellowships scheme, EPSRC/MURI grant EP/N019474/1 and FiveAI.
%, to Dr. Xiang Bai by the National Program for Support of Top-notch Young Professionals and the Program for HUST Academic Frontier Youth Team 2017QYTD08.

%This work was supported by National Key R&D Program of China (No. 2018YFB1004600), to Dr. Xiang Bai by the National Program for Support of Top-notch Young Professionals and the Program for HUST Academic Frontier Youth Team 2017QYTD08.

\newpage
\appendix
\section{Appendix}
In this supplementary material, we present more experimental details.

\vspace{1ex} \noindent \textbf{Video Feature Extraction.}~The interval for feature extraction is $8$. Layers after the global pooling layer of I3D~\cite{qiu2017learning} are discarded during feature extraction. The I3D network is pre-trained on Kinetics-400~\cite{carreira2017quo}.
Since many categories on \dbname{} are unique to MUSES, we finetune the I3D network on \dbname{}. Besides, optical flow is not used for feature extraction due to excessive computation cost.
On THUMOS14, we adopt the two-stream strategy~\cite{simonyan2014two} and extract features from both RGB and optical flow frames with two-stream I3D models pre-trained on Kinetics-400.
It's worth noting that the categories on THUMOS14 highly correlate with those on Kinetics-400. On ActivityNet-1.3, we employ the two-stream networks trained on ActivityNet-1.3 by Xiong \etal~\cite{xiong2016cuhk} for features extraction.

\vspace{1ex} \noindent\textbf{Proposal Generation.}~On THUMOS14, we use proposals generated by~\cite{lin2018bsn} for its excellent performance. On \dbname{}, we find it predicts too many false boundaries and achieves low recall rates, probability due to the difficulty in detecting event boundaries in multi-shot scenarios. Therefore, a different proposal generation method is employed. Following~\cite{shou2016temporal}, we generate sliding windows proposals of multiple lengths and employ a binary classifier to rank the proposals. The window is slid with a stride of 25\% of its length and the lengths are 10, 25, 40, 55, 70, 85, 100, 130, 160, and 190 seconds. The binary classifier is composed of a convolutional stage and a fully connected stage. The convolutional stage stacks 4 1D convolutional layers with 128, 256, 512, and 1024 filters of kernel size 3 respectively, each followed by a ReLU layer and a max-pooling layer with kernel size 2. The fully connected stage includes 2 fully connected layers of 512 and 3 units respectively.
After proposal ranking and non-maximal suppression (NMS) with a threshold $0.8$, the top 100 scored proposals are kept for the event localization model. 

\vspace{1ex} \noindent\textbf{Network Architecture.}~By default, we use two multi-scale blocks for Temporal Aggregation. 
In each multi-scale block, there are $K=4$ branches. The kernel sizes of these branches are $\{1 \times 3, 3 \times 3, 3 \times 3, 3 \times 3 \}$ and the corresponding unit sizes ($W$) are $\{3, 3, 6, 9\}$ respectively. 
The output channels of the first and the second block are 384 and 512 respectively.

For proposal feature extraction, we follow ~\cite{dai2017temporal, chao2018rethinking, zeng2019graph} to extend the boundaries of each proposal by 50\% of its length on both the left and right sides before RoI Pooling~\cite{girshick2015fast}, which is helpful for exploiting contextual information.

\vspace{1ex} \noindent\textbf{Loss Function.} For proposal classification, completeness classification and boundary regression, the cross-entropy loss, the hinge loss and the Smooth L1 loss are used respectively. The total loss is the weighted sum of the three losses, with weights of 1, 0.5, and 0.5 respectively. 

\vspace{1ex} \noindent\textbf{Training and Inference.}~During training, we set the initial learning rate to 0.01, mini-batch size to 32 and train the models with SGD optimizer with momentum 0.9. The models are trained for 20 epochs on THUMOS14 and 30 epochs on \dbname{}. After training for 15 epochs, the learning rate is divided by $10$. For post-processing, we apply NMS with a threshold $0.4$ to remove redundant detections. On THUMOS14, the predictions of the RGB and flow streams are fused using a ratio of $1:1.2$ during inference. On ActivityNet, the training and inference details are the same as~\cite{lin2019bmn}.

{\small
\bibliographystyle{ieee_fullname}
\bibliography{egbib}
}

\end{document}